# Chapter 15

# Boosting the Efficiency of Metaheuristics through Opposition-Based Learning in Optimum Locating of Control Systems in Tall Buildings

Salar Farahmand-Tabar[1*] · Sina Shirgir[2]
[1]Department of Civil Engineering Eng., Faculty of Engineering, University of Zanjan, Zanjan, Iran
*farahmandsalar@znu.ac.ir; farahmandsalar@gmail.com

[2]Department of Structural Engineering Eng., Faculty of Civil Engineering, University of Tabriz, Iran
 s.shirgir@tabrizu.ac.ir

**Abstract.** Opposition-based learning (OBL) is an effective approach to improve the performance of metaheuristic optimization algorithms, which are commonly used for solving complex engineering problems. This chapter provides a comprehensive review of the literature on the use of opposition strategies in metaheuristic optimization algorithms, discussing the benefits and limitations of this approach. An overview of the opposition strategy concept, its various implementations, and its impact on the performance of metaheuristic algorithms are presented. Furthermore, case studies on the application of opposition strategies in engineering problems are provided, including the optimum locating of control systems in tall building. A shear frame with Magnetorheological (MR) fluid damper is considered as a case study. The results demonstrate that the incorporation of opposition strategies in metaheuristic algorithms significantly enhances the quality and speed of the optimization process. This chapter aims to provide a clear understanding of the opposition strategy in metaheuristic optimization algorithms and its engineering applications, with the ultimate goal of facilitating its adoption in real-world engineering problems.

## 1. Introduction

Metaheuristic algorithms are powerful optimization techniques used to solve complex engineering problems that are difficult or impossible to solve using traditional optimization methods. These algorithms are based on various mathematical and computational concepts and are designed to explore the search space efficiently, identify promising solutions, and converge to the optimal solution. However, despite their effectiveness, metaheuristic algorithms have limitations in terms of efficiency, accuracy, and scalability. For example, some metaheuristic algorithms may converge slowly or get stuck in local optima, while others may need numerous fitness evaluations to achieve the optimum solution. Additionally, as the complexity of the problem increases, the performance of metaheuristic algorithms may degrade, making them less effective in solving real-world engineering problems.

Therefore, there is a need to improve the performance of metaheuristic algorithms to enhance their efficiency and effectiveness in solving complex engineering problems. This can be achieved by developing new and innovative algorithms that address the limitations of existing algorithms or by integrating various techniques, such as machine learning, opposition-based learning, and adaptive parameter control, into existing algorithms to enhance their performance. Moreover, improving the performance of metaheuristic algorithms can have significant implications for various engineering fields, such as mechanical, civil, and electrical engineering, where optimization plays a critical role in solving complex problems. It can lead to the development of more efficient and reliable systems, reduce costs, and improve productivity and safety [1-3]. Thus, research in this area is essential to address the challenges of the modern world and enhance the quality of life.

Opposition-Based Learning (OBL) is a technique used in optimization algorithms to generate opposite solutions to the current population. In OBL, each individual solution in the population is paired with its opposite solution, which is created by changing the sign of each decision variable's value. The opposite solutions are then evaluated and added to the population if they improve the overall population's diversity or fitness. The idea behind OBL is to increase the population's diversity and prevent the optimization algorithm from being trapped in local optima. By generating opposite solutions, OBL can explore the search space in a more efficient way, leading to better convergence and better-quality solutions. The OBL is a powerful technique that can enhance the efficiency and effectiveness of optimization algorithms by improving population diversity and exploring the search space more efficiently.

There are many literatures exist which used OBL [4] to improve different optimization algorithms and this issue have been proved that using OBL can boost the performance of the algorithms. For example, Sarkhel et al. [5] employed OBL in conjunction with the Harmony Search algorithm, aiming to enhance its convergence speed. Similarly, Shan et al. (2016) [6] utilized OBL to improve both the population diversity and convergence speed of the Bat Algorithm. In a different context, Sapre and Mini (2019) [7] employed OBL to enhance the convergence rate of the Moth Flame Optimization algorithm. Additionally, Zhou et al. (2017) [8] incorporated OBL into the memetic algorithm to enhance both its convergence speed and population diversity.

The protecting devices which are known as control devices, usually categorized into passive, active, and semi-active, according to the level and mechanisms of energy they require [9-13]. Semi-active controllable devices utilizing magnetorheological (MR) fluid have gained significant attention in various fields such as transportation vehicles, building suspensions, and biomedical applications over the past two decades due to their unique advantages. MR fluids consist of magnetically polarizable particles with sizes in the range of a few microns, dispersed within a carrying liquid such as mineral or silicon oil [14]. Its remarkable properties enable it to swiftly respond to magnetic field variations within milliseconds, while remaining unaffected by other factors that may disrupt its rheological characteristics [15-18]. MR actuators capitalize on the unique rheological behavior of MR fluids, offering advantages such as continuous adjustment, compact size, low energy consumption, and convenient control [19, 20]. As a result, these actuators find extensive utility in various domains associated with shock absorption and buffer engineering, encompassing automobiles [21, 22], bridges [23, 24], ships [25], and military applications [26].

The most research in the literature is related to the tuned mass dampers [27-35]. In recent years, several studies have focused on the modeling and control strategies for MR devices [36, 37]. However, there is a growing need for design methods that can reduce costs, manufacturing time, and enhance performance. The development of an optimal MR damper design involves considering numerous factors, which poses challenges when using traditional optimization methods.

The chapter can be divided into two main perspectives. First, the optimization algorithms which suffer from problems such as trapping in local optima due to lack of population diversity are chosen to be improved. To enhance their exploration and exploitation capabilities, opposition-based learning is added to the standard versions of the algorithms, creating an enhanced algorithm. Then, the proposed algorithms are applied to solve the optimum distribution of the structural control system using the Magnetorheological (MR) fluid damper. Since the optimal placement of MR damper is crucial for the structural performance, enhanced methods are better to be employed for optimum design. To evaluate the performance, a benchmark 40-story tall shear frame is selected and the results are compared considering different optimization methods.

## 2.  Enhancing Methods for Population Diversity

To enhance the performance of optimization algorithms, particularly in terms of population diversity, both methods of Opposition-Based Learning (OBL) is used. The OBL involves the use of opposition pairs to enhance the diversity of the population. It creates a new individual by generating a mirror image of an existing individual with respect to the midpoint of the search space. By incorporating the opposite characteristics, OBL helps to increase the diversity of the population, preventing the algorithm from getting stuck in local optima and improving the algorithm's exploration ability. By combining this method with standard optimization algorithms, the performance of the algorithms can be enhanced, resulting in faster convergence rates, better population diversity, and more efficient exploration and exploitation abilities. These techniques are particularly useful when dealing with complex optimization problems that require a good balance between exploration and exploitation, such as engineering design problems.

### 2.1.  Opposition-Based Learning

The first introduction of the opposition-based learning method was proposed by Tizhoosh [38] in the field of machine intelligence. The method is centered on the generation of an opposite number, which involves the production of a solution that is opposite to the current solution while being bounded within upper and lower limits. In other words, given a real number $h$ that is confined within the range of $a$ and $b$ ($h \in [a, b]$), its opposite can be obtained using the equation below.

$$\bar{h} = b + a - h \tag{1}$$

To expand this definition in $n$-dimensions, the following equation is utilized:

$$\bar{h}_i = b_i + a_i - h_i, \qquad i = 1, 2, \dots, N \tag{3}$$

The opposite vector $\bar{h} \in R^n$ is derived from the real vector $h \in R^n$. The fitness function is used to compare the two solutions ($h$ and $\bar{h}$), and the better solution is stored. For instance, $\bar{h}$ is saved if $f(h) \leq f(\bar{h})$ (for maximization), otherwise $h$ is stored. The ability of OBL to improve the convergence rate of an optimizer has recently gained significant attention. Specifically, it doubles the population in each iteration to achieve a better population. Within each iteration, a fresh initial population is generated, and the superior individuals, constituting half of the population, are retained, while the remaining individuals are discarded. This characteristic significantly enhances the convergence rate of the optimization algorithm.

## 2.2. Enhancing Methods by Population Diversity

One of the key challenges in metaheuristic optimization is maintaining a diverse population of candidate solutions throughout the search process, as this can help to avoid premature convergence and increase the possibility of finding high-quality solutions. In recent years, there has been a growing interest in developing techniques to enhance the search strategy in metaheuristics, with a particular focus on improving the efficiency and effectiveness of these methods.

### 2.2.1. Gravitational Search Algorithm

The gravitational search algorithm (GSA) is a stochastic search algorithm inspired by nature that is commonly employed for optimization problems. It was first introduced by E. Rashedi [39] with the aim of addressing non-linear optimization issues. A comprehensive investigation of GSA and a brief review of its developments in solving various engineering problems were discussed in [40], aiming to build up a global picture to explore possible applications. This algorithm is according to Newton's theory of gravity, which posits that particles attract each other via a gravitational force. The force exerted between two particles is directly related to the product of their masses and inversely related to the square of the distance between them. In the proposed algorithm, particles are treated as entities whose per-formance is assessed based on their respective masses. The GSA assigns four characteristics to each particle, namely its position, inertial mass, active and passive gravitational mass. The particle's position represents the problem solution, while the gravitational and inertial masses are calculated using the fitness function.

Like all population-based algorithms, the GSA has the ability to explore and exploit the search space. It employs exploration at the beginning to avoid getting trapped in local optima and switches to exploitation later on. A time function, known as the K-best particle/agent, is employed to exert attraction on other particles, thereby enhancing the performance of the Gravitational Search Algorithm (GSA). This approach facilitates a balance between exploration, aimed at discovering new solutions, and exploitation, focused on refining promising solutions. The value of the K-best function gradually decreases with time, ultimately leaving only one particle with a heavy mass that denoted the final solution. The procedure of the GSA is detailed step-by-step as follows:

***Step 1:*** Generate initial population ($S$) and computing the fitness of each agent.

***Step 2:*** Update the inertial mass $Mi(t)$, the best agent best($t$), the worst agent worst($t$), and the center of mass $G(t)$ for each agent using the relevant equations:

$$M_i(t) = \frac{m_i(t)}{\sum_1^N m_i(t)}, \qquad m_i(t) = \frac{fit_i(t) - worst(t)}{best(t) - worst(t)}$$

$$best(t) = \min_{j\in\{1,...,n\}} fit_j(t), \quad best(t) = \max_{j\in\{1,...,n\}} fit_j(t) \tag{3}$$

$$G(t) = G_0 \times e^{-\tau\left(\frac{iter}{iter_{max}}\right)}, \qquad G_0 = 100 \ \& \ \tau = 20$$

***Step 3:*** Calculate the total forces acting on each agent in different directions:

$$F_i^d(t) = \sum_{j\in Kbest, j\neq i}^{N} rand_j \, G(t) \times \frac{M_i(t) \times M_j(t)}{R_{ij}(t) + \varepsilon} \times \left(x_j^d(t) - x_i^d(t)\right) \tag{3}$$

***Step 4:*** Calculate the acceleration of each agent, and their velocity:

$$a_i^d(t) = \frac{F_i^d(t)}{M_i(t)} = \sum_{j\in Kbest, j\neq i}^{N} rand_j \, G(t) \times \frac{M_j(t)}{R_{ij}(t) + \varepsilon} \times \left(x_j^d(t) - x_i^d(t)\right) \tag{3}$$

$$V_i^d(\mathrm{t}+1) = rand_i \times V_i^d(\mathrm{t}) + a_i^d(\mathrm{t})$$

***Step 5:*** Update the positions of each agent:

$$x_i^d(\text{t}+1) = x_i^d(\text{t}) \times V_i^d(\text{t}+1) \tag{3}$$

***Step 6:*** Check for any constraints on the problem.
***Step 7:*** Repeat steps 2 to 6 until the termination criterion is satisfied.

## 2.2.2. Big Bang- Big Crunch

The Big Bang- Big Crunch (BB-BC) algorithm [41] is according to the Big Bang and Big Crunch theory and is computationally efficient with acceptable convergence speed. The BB-BC exhibits diverse adaptations, and it has found particular utility in engineering optimization specially the power systems. Its versatility and efficacy in addressing various challenges within this domain have led researchers to explore and develop different variations of the algorithm for improved performance. [42]. It has two stages: Big Bang generates random candidate solutions, while Big Crunch orders them based on quality. In each iteration, a new population is generated around the center of mass calculated during Big Crunch. After multiple iterations, randomness decreases, and the algorithm converges to a solution. After the Big Bang step in the BB-BC algorithm, the Big Crunch step is executed as a convergence operator. The output of the Big Crunch step is calculated by finding the center of mass using the inverse fitness function values as input. The resulting point, which serves as the center of mass, is denoted as $x^c$. It is determined using the following calculation:

$$x^c = \frac{\sum_{i=1}^{N} \frac{1}{f^i} x^i}{\sum_{i=1}^{N} \frac{1}{f^i}} \tag{3}$$

The variable $x^i$ represents a point randomly generated within a search space with $n$ dimensions, while $f^i$ is the corresponding fitness value of that point. The population size of the algorithm during the Big Bang step is represented by the variable $N$. Once the Big Crunch stage is complete, the algorithm needs to generate new candidates to be utilized as the starting population for the next Big Bang stage. This process of generating new members is essential for the algorithm's continued optimization:

$$x_i^{new} = X^c + \frac{r\alpha(x_{max} - x_{min})}{iter} \tag{3}$$

The parameter $\alpha$ regulates the size of the search space, controlling the extent of exploration. On the other hand, $r$ denotes a random number generated from a standard normal distribution, and its value varies for each can-didate solution. The optimization problem variables have upper and lower limits, represented by xmax and xmin, respectively. After the second explosion, the center of mass is recomputed. The steps of explosion and contraction are repeated until a termination criterion is satisfied.

To summarize the steps in the BB-BC algorithm:

***Step 1:*** Generate $N$ candidate solutions randomly within the search space limits.

***Step 2:*** Compute the values of fitness function for all candidate solutions.

***Step 3:*** Determine the center of mass, or the best-fit individual can be used instead.

***Step 4:*** Calculate new candidate solutions around the center of mass by subtracting or adding a normal random number that decreases in value as iterations progress.

***Step 5:*** Return to Step 2 until the termination criteria are satisfied.

### 2.2.3. Particle Swarm Optimization

The Particle Swarm Optimization (PSO) algorithm [43] is enhanced by a swarm of particles that continuously adjust their positions from one iteration to the next in order to optimize the search process. Considering the hybridization, enhancement, and various variants of the PSO, its real-world applications are categorized into several areas, such as health-care, environmental, industrial, commercial, smart city, and general aspects applications [44]. To achieve the best possible solution in PSO, each particle moves towards its own best previous position ($p_{\text{best}}$) as well as the overall best position ($g_{\text{best}}$) within the swarm. When the objective of the problem is minimization, the following equation holds true:

$$p_{best_i}^t = x_i^* \mid \text{f}(x_i^*) = \min_{\substack{k=\{1,2,\ldots,t\} \\ i=\{1,2,\ldots,N\}}} (\{\text{f}(x_i^k)\})$$

$$g_{best_i}^t = x_*^t \mid \text{f}(x_*^t) = \min_{\substack{i=\{1,2,\ldots,N\} \\ k=\{1,2,\ldots,t\}}} (\{\text{f}(x_i^k)\})$$

(3)

In the given equation, $i$ corresponds to the index of the particle, $t$ indicates the current iteration number, $f$ denotes the objective function that is being optimized or minimized, $x$ represents the position vector of the particle (or a potential solution), and $N$ signifies the total number of particles in the swarm. The velocity ($v$) and position ($x$) of each particle $i$ are updated using the following equations at each iteration $t + 1$:

$$V_i^{t+1} = \omega V_i^{t+1} + c_1 r_1 (p_{best_i}^t - x_i^t) c_2 r_2 (g_{best_i}^t - x_i^t)$$

(3)

$$x_i^{t+1} = x_i^t + V_i^{t+1}$$

The velocity vector ($V$) is used in the update equations and is dependent on the inertia weight ($\omega$), which balances the exploitation of local search and the exploration of global search. Random vectors $r_1$ and $r_2$ are uniformly distributed within the range of $[0,1]^D$, where $D$ is the dimensionality of the search space or the size of the problem being solved. The positive constants $c_1$ and $c_2$, known as acceleration coefficients, are also utilized in the equations.

The PSO algorithm utilizes the following steps to find the best solution:

***Step 1:*** The initial vectors of position and velocity for each particle are generated randomly.

***Step 2:*** The objective function is evaluated with respect to the predefined constraints.

***Step 3:*** The value fitness for each particle is compared to its previous best. If the current value is better, it replaces the previous value.

***Step 4:*** The value of fitness related to each particle is also compared to the global best. If the current value is better, it is replaced as the new global best.

***Step 5:*** The velocity and position vectors of each particle are updated.

***Step 6:*** Steps (2) through (5) are repeated until the termination criteria are satisfied or the predefined maximum number of iterations is reached.

During each iteration, the maximum velocity and position are predetermined, and they control the changes in velocity and position of each particle.

## 3. Outline for Optimum Locating of MR dampers

This section proposes an optimum design of semi-active control system for a tall shear building based on locating of MR dampers using the metaheuristic algorithms. A 40-story shear building is developed to investigate the behavior of control systems. The structural properties such as mass, damping, stiffness, elevation, and moment of inertia are shown in Table 1. The distribution of MR dampers including location and number of dampers in each story are chosen as design variables. The sum of the ratio of maximum controlled drift to uncontrolled drift of all stories taken as the target of optimization so that the objective function can be evaluated as:

$$Objective\ function = \sum_{i=1}^{n} \frac{drift_{i_{Controlled}}^{Max}}{drift_{i_{Uncontrolled}}^{Max}} \tag{16}$$

where $n$ represents the number of stories.

The constraints of the present optimization problem are defined for the number of dampers in each story. In this regard, the allowable value is defined between 0 to 5, in which 0 represents the unuse of damper and the 5 is related to the maximum number of dampers that can be installed in stories. One of the other constraints is the total number of dampers installed in the building structure, which is selected as 40 dampers according to the story number of the structural model. The maximum number of iterations is selected as the stopping criteria of the algorithms equal to 500. The statistical results are based on 30 independent experiments for a reliable assessment, and the design results are reported based on the best solution.

The structural model has been analyzed in both controlled and uncontrolled cases using a numerical time-history method. The analysis was conducted considering a white noise ground acceleration with a peak ground acceleration ($PGA$) of 0.4 g. This ground acceleration, denoted as $W(t)$, is shown in Figure 1. The selected record serves as an example to demonstrate the methodology. Also, to examine the effect of the distribution pattern of MR dampers on the behavior of tall buildings, seismic analysis under historical earthquakes is considered. In this regard, the records of some earthquakes are selected, and the details of these events are tabulated in Table 2.

**Table 1** The story-related parameters of the 40-story frame.

| Story | $m_i(t)$ | $k_i(MN/m)$ | $c_i(MNs/m)$ | $I_i(kgm^2)$ | $z_i(m)$ |
|-------|----------|-------------|--------------|--------------|----------|
| Base  | 1960     | -           | -            | $1.96 \times 10^8$ | 0 |
| 1     | 980      | 2130.00     | 42.60        | $1.31 \times 10^8$ | 4 |
| 2     | 980      | 2100.97     | 42.02        | $1.31 \times 10^8$ | 8 |
| 3     | 980      | 2071.95     | 41.44        | $1.31 \times 10^8$ | 12 |
| 4     | 980      | 2042.92     | 40.86        | $1.31 \times 10^8$ | 16 |
| 5     | 980      | 2013.90     | 40.28        | $1.31 \times 10^8$ | 20 |
| 6     | 980      | 1984.87     | 39.70        | $1.31 \times 10^8$ | 24 |
| 7     | 980      | 1955.85     | 39.12        | $1.31 \times 10^8$ | 28 |
| 8     | 980      | 1926.82     | 38.54        | $1.31 \times 10^8$ | 32 |
| 9     | 980      | 1897.79     | 37.96        | $1.31 \times 10^8$ | 36 |
| 10    | 980      | 1868.77     | 37.38        | $1.31 \times 10^8$ | 40 |
| 11    | 980      | 1839.74     | 36.79        | $1.31 \times 10^8$ | 44 |
| 12    | 980      | 1810.72     | 36.21        | $1.31 \times 10^8$ | 48 |
| 13    | 980      | 1781.69     | 35.63        | $1.31 \times 10^8$ | 52 |
| 14    | 980      | 1752.67     | 35.05        | $1.31 \times 10^8$ | 56 |

| 15 | 980 | 1723.64 | 34.47 | $1.31 \times 10^8$ | 60 |
|----|-----|---------|-------|--------------------|-----|
| 16 | 980 | 1694.62 | 33.89 | $1.31 \times 10^8$ | 64 |
| 17 | 980 | 1665.59 | 33.31 | $1.31 \times 10^8$ | 68 |
| 18 | 980 | 1636.56 | 32.73 | $1.31 \times 10^8$ | 72 |
| 19 | 980 | 1607.54 | 32.15 | $1.31 \times 10^8$ | 76 |
| 20 | 980 | 1578.51 | 31.57 | $1.31 \times 10^8$ | 80 |
| 21 | 980 | 1549.49 | 30.99 | $1.31 \times 10^8$ | 84 |
| 22 | 980 | 1520.46 | 30.41 | $1.31 \times 10^8$ | 88 |
| 23 | 980 | 1491.44 | 29.83 | $1.31 \times 10^8$ | 92 |
| 24 | 980 | 1462.41 | 29.25 | $1.31 \times 10^8$ | 96 |
| 25 | 980 | 1433.38 | 28.67 | $1.31 \times 10^8$ | 100 |
| 26 | 980 | 1404.36 | 28.09 | $1.31 \times 10^8$ | 104 |
| 27 | 980 | 1375.33 | 27.51 | $1.31 \times 10^8$ | 108 |
| 28 | 980 | 1346.31 | 26.93 | $1.31 \times 10^8$ | 112 |
| 29 | 980 | 1317.28 | 26.35 | $1.31 \times 10^8$ | 116 |
| 30 | 980 | 1288.26 | 25.77 | $1.31 \times 10^8$ | 120 |
| 31 | 980 | 1259.23 | 25.18 | $1.31 \times 10^8$ | 124 |
| 32 | 980 | 1230.21 | 24.60 | $1.31 \times 10^8$ | 128 |
| 33 | 980 | 1201.18 | 24.02 | $1.31 \times 10^8$ | 132 |
| 34 | 980 | 1172.15 | 23.44 | $1.31 \times 10^8$ | 136 |
| 35 | 980 | 1143.13 | 22.86 | $1.31 \times 10^8$ | 140 |
| 36 | 980 | 1114.10 | 22.28 | $1.31 \times 10^8$ | 144 |
| 37 | 980 | 1085.08 | 21.70 | $1.31 \times 10^8$ | 148 |
| 38 | 980 | 1056.05 | 21.12 | $1.31 \times 10^8$ | 152 |
| 39 | 980 | 1027.03 | 20.54 | $1.31 \times 10^8$ | 156 |
| 40 | 980 | 998.00  | 19.96 | $1.31 \times 10^8$ | 160 |

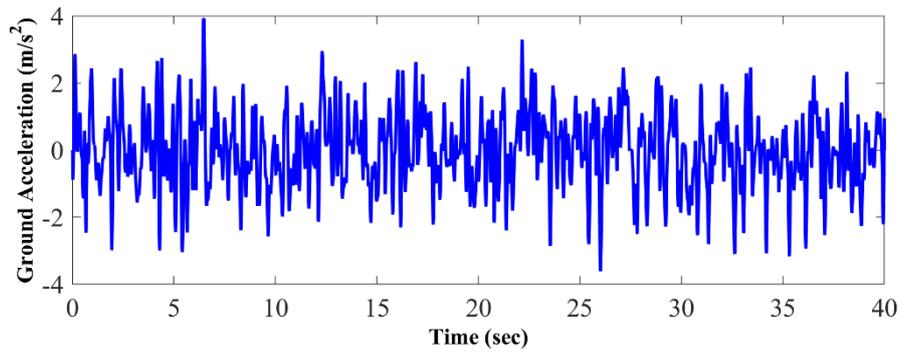

**Fig. 1.** Time-history of white noise ground acceleration, $W(t)$, $PGA = 0.4$g.

**Table 2** Characteristic data of ground motions

| No. | Name | Recording station | Year | Magnitude | FN component |
|-----|------|-------------------|------|-----------|--------------|
| 1 | Northridge | Beverly Hills - Mulhol | 1994 | 6.7 | NORTHR/MUL009 |
| 2 | Kobe | Shin-Osaka | 1995 | 6.9 | KOBE/SHI000 |
| 3 | Landers | Coolwater | 1992 | 7.3 | LANDERS/CLW-LN |

The evaluation of control performance involves the use of several performance indices that compare the uncontrolled and controlled responses of the structure. Six criteria are employed, where the first half represent the maximum values, and the remaining are the structural norm response. These criteria are summarized in Table 3. The criteria consist of the following:

- o  $\delta^{max}$: Max. un-controlled inter-story drift.
- o  $\ddot{x}_u^{max}$: Absolute value of the roof acceleration.
- o  $F_b^{max}$: Max. force of the base shear.
- o  $d_i(t)$: Controlled inter-story drift of the $i$th level.
- o  $\ddot{x}_{ai}(t)$: Absolute acceleration of the $i$th level.
- o  $m_i$: Seismic mass of the $i$th level.

Additionally, the normed operator $\|.\|$ is applied to calculate the norm values of the responses.

**Table 3** Performance criteria for the building with the control system

- Peak inter-story drift:

$$J_1 = max \left\{ \frac{\underset{t.i}{max} \left( |d_i(t)| \right)}{\delta^{max}} \right\}$$

- Peak level acceleration:

$$J_2 = max \left\{ \frac{\underset{t.i}{max} \left\{ \ddot{x}_{ai}(t) \right\}}{\ddot{x}_u^{max}} \right\}$$

- Peak base shear force:

$$J_3 = max \left\{ \frac{\underset{t}{max} \left| \sum_i m_i \ddot{x}_{ai}(t) \right|}{F_b^{max}} \right\}$$

- Normed inter-story drift:

$$J_4 = max \left\{ \frac{\underset{t.i}{max} \left( \|d_i(t)\| \right)}{\|\delta^{max}\|} \right\}$$

- Normed level acceleration:

$$J_5 = max \left\{ \frac{\underset{t.i}{max} \|\ddot{x}_{ai}(t)\|}{\|\ddot{x}_u^{max}\|} \right\}$$

- Normed base shear force:

$$J_6 = max \left\{ \frac{\underset{t}{max} \left\| \sum_i m_i \ddot{x}_{ai}(t) \right\|}{\|F_b^{max}\|} \right\}$$

The flowchart of the proposed method is depicted in Fig. 2.

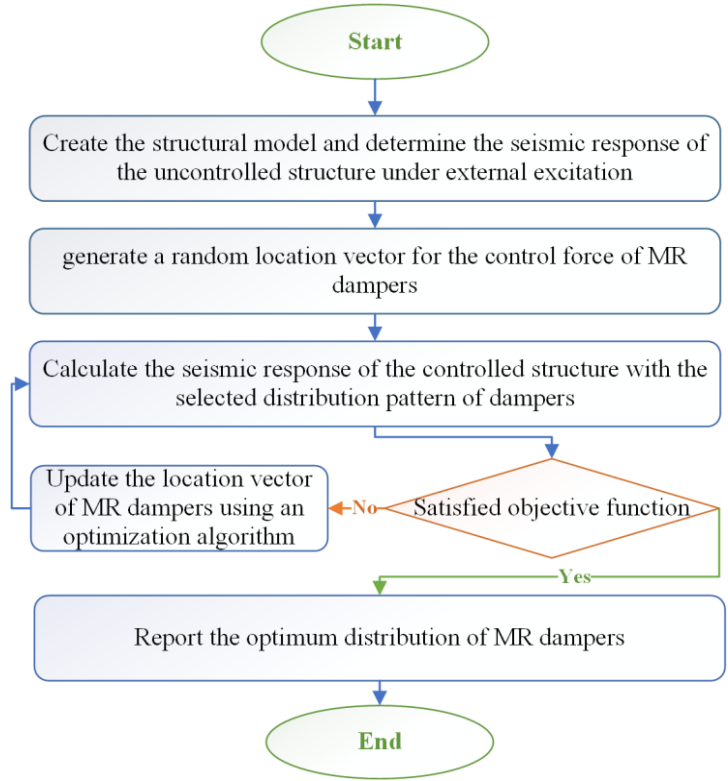

**Fig. 2** Flowchart for the optimum locating of the MR damper

## 4. Results and Discussion on the case study

In this section, the efficiency of the proposed method to enhance the optimization algorithms is discussed. The results of the numerical study are presented for the optimum distribution of MR dampers in a tall building. Fig. 3 shows the convergence history of the objective function defined in previous section. According to this figure, the performance of the algorithms using opposition-based learning technique are better compared to their standard version. This enhancement is visually evident in the faster convergence and reduced fluctuations observed in the convergence history. When comparing the considered methods, it is worth noting that PSO with the value of 32.6064 emerges as the most effective among the alternatives, outperforming BB-BC ($f_{best}$=32.8892) and GSA ($f_{best}$=33.5686) in terms of convergence speed and overall optimization performance. The conver-gence curve for PSO in Fig. 3 demonstrates a steeper descent, indicating its capability to converge to an optimal solution quicker (within 308 iterations).

The quantitative results further support these observations, as displayed in Table 4. The table provides statistical data, including the best and average results obtained from all independent runs, the corresponding standard deviations, and the iteration count for achieving convergence. As expected, the OBL technique proves to be a valuable asset for improving the performance (approximately 5% in average) of all optimization methods considered in the study. The lower standard deviation achieved using OBL technique reveals the higher reliability of the obtained solutions in comparison with those from standard algorithms.

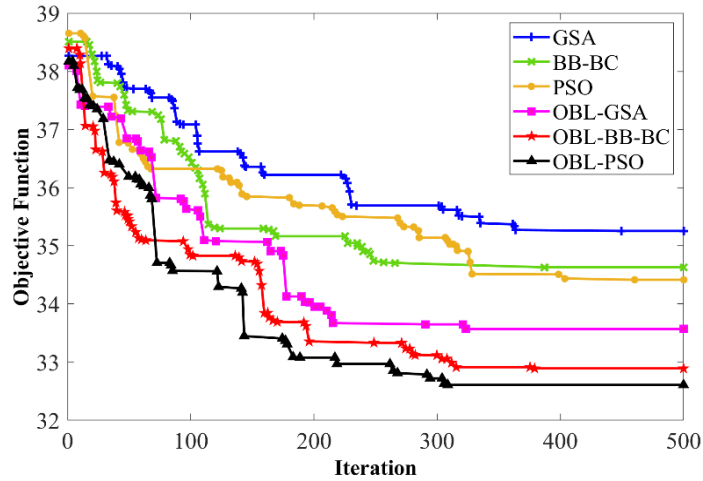

**Fig. 3** Comparative convergence history of the objective function

**Table 4** Statistical analysis of optimum results

| Case | PSO | OBL-PSO | BB-BC | OBL-BB-BC | GSA | OBL-GSA |
|---|---|---|---|---|---|---|
| $f_{best}$ | 34.4143 | 32.6064 | 34.6299 | 32.8892 | 35.2556 | 33.5686 |
| $Diff.$ (%) | | 5.25 | | 5.03 | | 4.79 |
| $f_{ave}$ | 34.9835 | 33. 0176 | 35.2567 | 33.3601 | 35.9845 | 34.0869 |
| $\sigma$ | 0.6779 | 0.5205 | 0.7388 | 0.5897 | 0.8366 | 0.6323 |
| $iter_{best}$ | 460 | 308 | 387 | 379 | 449 | 383 |

Table 5 contains the seismic response of structure for different distribution patterns of dampers under various seismic ground motions. The performance criteria ($J_1$ -$J_6$) are based on Table 3. According to Table 5, the results achieved by OBL version of the algorithms are better than those of standard version. Among these methods, PSO outperforms its rivals both in standard and OBL version. Based on the obtained values of $J$ indices, distribution patterns of the dampers represent the efficient performance of the system.

**Table 5** Seismic response of structure for different distribution patterns of dampers under various events

| EQ event | Index | Optimization Algorithm | | | | | |
|---|---|---|---|---|---|---|---|
| | | PSO | OPSO | BB-BC | OBB-BC | GSA | OGSA |
| Northridge | $J_1$ | 0.8154 | **0.7941** | 0.8714 | 0.8379 | 0.8396 | 0.8267 |
| | $J_2$ | 0.7254 | **0.6984** | 0.7497 | 0.7147 | 0.7467 | 0.7239 |
| | $J_3$ | 0.7002 | **0.6827** | 0.7514 | 0.7364 | 0.8124 | 0.8073 |
| | $J_4$ | 0.7598 | 0.7341 | 0.7315 | **0.7045** | 0.7287 | 0.7354 |
| | $J_5$ | 0.6547 | 0.6521 | 0.6669 | **0.6396** | 0.6784 | 0.6659 |
| | $J_6$ | 0.6854 | **0.6587** | 0.7358 | 0.7149 | 0.7496 | 0.7236 |
| Kobe | $J_1$ | 0.8954 | **0.8031** | 0.9145 | 0.8274 | 0.9283 | 0.8313 |
| | $J_2$ | 0.8475 | **0.8156** | 0.9358 | 0.8846 | 0.9045 | 0.8714 |
| | $J_3$ | **0.8547** | 0.8829 | 0.8976 | 0.9001 | 0.9385 | 0.9209 |
| | $J_4$ | 0.8969 | **0.8547** | 0.8974 | 0.8661 | 0.90012 | 0.8824 |
| | $J_5$ | 0.8415 | **0.7951** | 0.8752 | 0.8462 | 0.8996 | 0.8553 |
| | $J_6$ | **0.8045** | 0.8374 | 0.9447 | 0.8941 | 0.9138 | 0.8997 |
| Landers | $J_1$ | 0.7858 | **0.7347** | 0.8219 | 0.8097 | 0.8665 | 0.8396 |
| | $J_2$ | 0.7146 | **0.6737** | 0.7521 | 0.6987 | 0.7986 | 0.7482 |
| | $J_3$ | 0.7359 | **0.7058** | 0.7637 | 0.7491 | 0.8056 | 0.7964 |
| | $J_4$ | 0.7834 | **0.7325** | 0.7956 | 0.7694 | 0.8319 | 0.8003 |
| | $J_5$ | 0.6825 | **0.6588** | 0.7167 | 0.6879 | 0.7425 | 0.7058 |
| | $J_6$ | 0.6728 | **0.6587** | 0.7296 | 0.7358 | 0.7348 | 0.7148 |

Based on the $J$ indices obtained from the analysis, the distribution patterns of dampers demonstrate the efficient performance of the structural system. Fig. 4 illustrates the distribution of the number and location of MR dampers across the stories of the tall 40-story frame. The figure reveals that certain stories do not require any dampers, while others necessitate multiple dampers. Although different methods propose varying distributions of dampers throughout the structure, there are specific stories where the highest number of dampers (more than 4) is consistently required across all methods. Notably, stories such as 5 and 26 exhibit optimum performance in all methods without the need for applying any dampers.

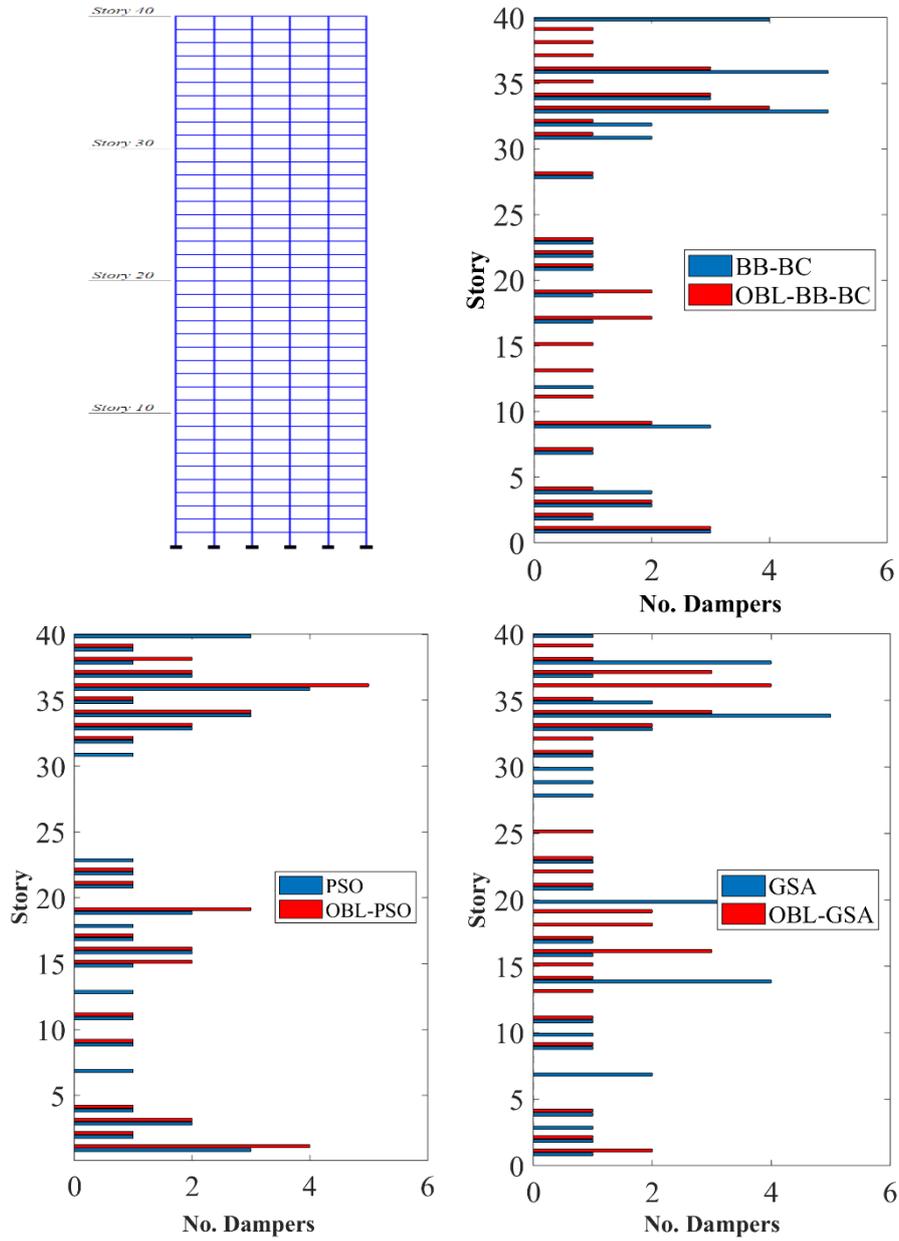

**Fig. 4** Number of dampers obtained for the optimum structure

The incorporation of the OBL technique in the optimization process leads to remarkable enhancements, reflected in both the visual convergence pat-terns and the statistical outcomes. These advancements contribute significantly to achieving an optimized distribution of MR dampers in tall buildings, promising increased structural stability and enhanced performance during seismic events. These results emphasize the significance of incorporating the opposition-based learning technique in optimization algorithms, providing a valuable tool for engineers and researchers seeking efficient and reliable solutions for complex structural optimization problems.

## 5. Conclusions

This chapter investigated the efficiency of the Opposition-Based Learning (OBL) algorithms dealing with the optimization of a structural problem. To overcome metaheuristics' problems such as low convergence rate and sticking in the local optima, OBL was added to several optimization algorithms to enhanced their exploration and exploitation capabilities. The physics-based algorithms such as Big Bang-Big Crunch (BB-BC), Particle Swarm Optimization (PSO), and Gravitational Search Algorithm (GSA) were chosen for this study. The performance of the proposed algorithms was examined considering a forty-story tall building frame equipped with MR damper to find the optimum number and location of the control devices. The optimum design criteria were to reduce the structural response subjected to earthquake excitations (Northridge, Kobe, Landers) by considering the optimum location of the MR dampers. Various structural responses such as maximum inter-story drift, story acceleration, base shear, and their norm values were considered. Based on the obtained values of $J$ indices, distribution patterns of the dampers denote the efficient performance of the system. According to results of the investigation, different distributed dampers can be achieved within the building height. The results were improved approximately 5% in average using the OBL technique considering their standard versions. Also, among the utilized methods, PSO outperforms its rivals in both standard OBL-based version.